\documentclass[runningheads]{llncs}

\usepackage[utf8]{inputenc}
\usepackage{graphicx}
\usepackage{dirtytalk}
\usepackage{multirow}
\usepackage{tikz}
\usepackage{tabularx}
\usepackage{adjustbox}
\usepackage{enumitem}
\usepackage{makecell}
\usepackage{subcaption}
\usepackage{caption} 
\usepackage{hyperref}
\usepackage{tabu}
\usepackage{caption}

\newcommand{\itelos}{\textit{iTelos}}

\title{Popularity Driven Data Integration}

\author{Fausto Giunchiglia\orcidID{0000-0002-5903-6150}\inst{1} \and
Simone Bocca\orcidID{0000-0002-5951-4589}\inst{1} \and Mattia Fumagalli\orcidID{0000-0003-3385-4769}\inst{2} \and Mayukh Bagchi \orcidID{0000-0002-2946-5018}\inst{1} \and Alessio Zamboni\orcidID{0000-0002-4435-1748}\inst{1}\thanks{The research by F. Giunchiglia, M. Bagchi and S. Bocca has received funding from the \emph{“DELPhi - DiscovEring Life Patterns”} project funded by the MIUR (PRIN) 2017. The research by A. Zamboni was supported by the \emph{InteropEHRate} project, EC Horizon 2020 programme under grant number 826106.}}

\institute{Department of Information Engineering and Computer Science (DISI),\\ University of Trento, Italy\\
\email{\{fausto.giunchiglia,simone.bocca,mayukh.bagchi,alessio.zamboni\}@unitn.it}
\and
Conceptual and Cognitive Modeling Research Group (CORE),\\
Free University of Bozen-Bolzano, Bolzano, Italy\\
\email{\{mattia.fumagalli@unibz.it\}}
}

\date{March 2021}

\begin{document}

\maketitle

\begin{abstract}
\vspace{-0.2cm}
\vspace{-0.2cm}
More and more, with the growing focus on large scale analytics, we are
 confronted with the need of integrating data from multiple sources. The problem is that these data are impossible to reuse \textit{as-is}.  The net result is high cost, with the further drawback that the resulting integrated data will again be hardly reusable \textit{as-is}. 
 \itelos\footnote{Not to be confused with \textit{Telos} \cite{mylopoulos1990telos}.} is a general purpose methodology aiming at minimizing the effects of this process. 
%
The intuition is that data will be treated differently based on their \textit{popularity}: the more a certain set of data have been reused, the more they will be reused and the less they will be changed across reuses, thus decreasing the overall data preprocessing costs, while increasing backward compatibility and future sharing.
%
\vspace{-0.2cm}

\end{abstract}

\keywords{Data Reuse \and Data Sharing \and Knowledge Graphs.}

\vspace{-0.1cm}
\section{Introduction} \label{sec1_introduction}

More and more, with the growing focus on large scale analytics, we are
 confronted with the need of integrating data from multiple sources.
%
The key issue is how to handle the \textit{semantic heterogeneity} which is intrinsic in such data. Two main approaches have been proposed. 
The first approach consists of using \textit{ontologies}, where the goal is to agree on a fixed language and/or schema towards facilitating future sharing \cite{2017-obdi-for-mdee}. 
The second consists of exploiting the flexibility of \textit{Knowledge Graphs (KGs)} \cite{2019-Kejriwal-KG}, as the means for facilitating the adaptation and integration of heterogeneous data.
However the problem is far from being solved. No matter the approach, it is just impossible to reuse data \textit{as-is}. 
The net result is usually a lot of data preprocessing (e.g., cleaning, normalization) which results in high cost, with the further drawback that the resulting integrated data will again be hardly reusable \textit{as-is}. It is a negative loop which consistently reinforces itself. 

 In this paper, we propose \itelos, a general purpose methodology 
 whose main goal is to minimize the high costs of this loop of reuse.
 \itelos\ is crucially based on the use of KGs and  ontologies, i.e., \textit{reference schemas}, as the schemas of KGs. However, its novel underlying intuition is to treat data differently, depending on their \textit{popularity}. In particular, the idea is to select first and minimize changes on those data which are more (re)-used thus decreasing the preprocessing costs, while increasing backward compatibility and future sharing. To this extent, \itelos\ distinguishes among three categories of data. That is: \textit{Common}, which are used across domains (e.g., data about space, time, transportation), \textit{Core}, which, while being more vertical than common, are extensively used in the domain under consideration (e.g., in tourism, all the data that can be found in Open Data portals),
 and \textit{Contextual}, namely, data specific to the application at hand (e.g., the data extracted on purpose from legacy systems). Popularity also drives the selection of the reference schemas, based on the intuition that, also at the schema level, more reuse is a good motivation for further reuse. Thus we have \textit{Common} reference schemas (e.g., standards about space, time, transportation, \textit{schema.org}\footnote{\url{https://schema.org/}}), \textit{Core} reference schemas,  (e.g., in Health, FHIR\footnote{\url{http://hl7.org/fhir/}}), and \textit{Contextual} reference schemas (e.g., as they apply the current application). 
 
\itelos\ implements the above idea based on a precisely articulated data integration process, based on three key assumptions, as follows:
 \begin{itemize}
 \vspace{-0.1cm}
  \item \textit{data} and  \textit{reference schemas} should be integrated  under the overall guidance of the needs to be satisfied, formalized as \textit{competence queries} (\textit{CQs}) \cite{gruninger1995role};
 \item The requirements, including those on data and reference schema reuse, as well as CQs, should be known a priori as part of an application \textit{purpose};
  \item  A difficulty is that the schemas of the data to be reused usually do not map to reference schemas, the mapping being usually arbitrarily complex. The idea is to build the integrated KG via a sequence of \textit{middle-out} iterations where, first, CQs are used to drive the selection and preprocessing of the data, largely independently from the reference schemas, and where, in a second step, reference schemas, suitably and independently integrated among them, are adapted to fit best the integrated data minimizing the negative effects on sharability.  
 \end{itemize}
 \noindent
This paper is organized as follows. 
In Section \ref{sec2-purpose} we describe the \itelos\ process. In Section \ref{sec5_reuse}  we describe how \itelos\ enhances the reusability, with minimal changes, of the available data. In Section
\ref{sec6_share} we describe how \itelos\ enhances the future sharability of the resulting KG. Finally, Section \ref{sec8_case} syntethically describes the case studies to which \itelos\ has been applied.



\section{The Process}
\label{sec2-purpose}

\begin{figure}[ht!]
\vspace{-0.6cm}
    \centering
    \makebox[\textwidth][c]{\includegraphics[width=15cm,height=3.8cm]{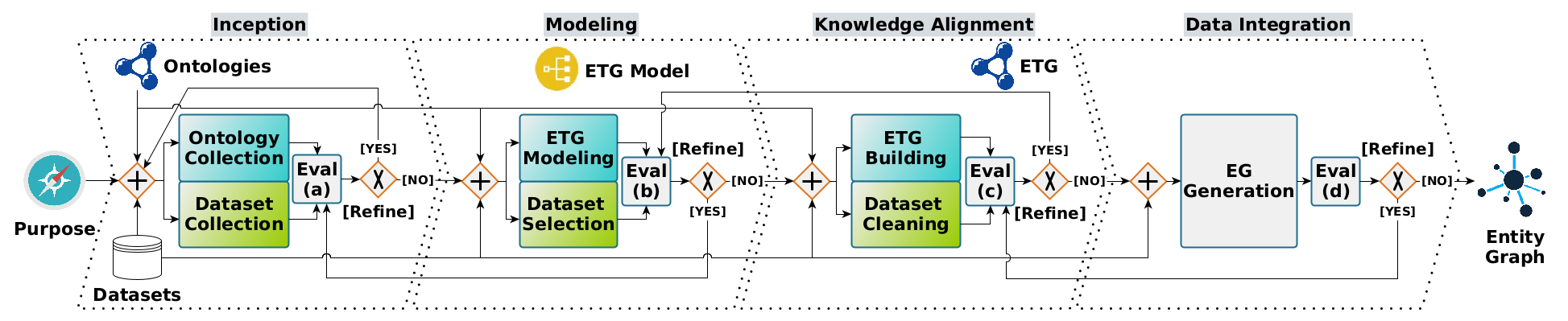}}
    \caption{The \itelos\ Process.}
    \label{fig:pipeline}
    \vspace{-0.7cm}
\end{figure}

The \itelos\ process is depicted in Figure \ref{fig:pipeline}. The \textit{User} provides in input the specification of the problem, the \textit{Purpose}, and receives in output an integrated KG, i.e., the \textit{Entity Graph}. 
The purpose contains three main elements, as follows:

\begin{itemize}
\vspace{-0.1cm}
    \item the functional requirements of the KG to be generated, that we assume to be ultimately formalized as CQs; 
    \item The \textit{datasets} to be reused. We assume that these datasets consist of \textit{Entity Graphs (EGs)}, namely graphs where nodes are \textit{entities} (e.g., my cat \textit{Garfield}), decorated with data property values and linked among them via object property links;\footnote{Many Open data portals provide such datasets, at different levels of formalization in the 5STAR Open Data Model.}
    \item The \textit{Ontologies} to be reused.  We assume that these ontologies, i.e., reference schemas, consist of \textit{entity type (etype) Graphs (ETGs)}, namely KGs which define the schema of EGs. In ETGs nodes are \textit{etypes}, namely classes of entities (e.g., the class \textit{cat)}, decorated by the data and object properties which define the EG structure.\footnote{Many such schema repositories are already available; see for instance: LOV (\url{https://lov.linkeddata.es/}), LOV4IoT (\url{http://lov4iot.appspot.com/}),  DATAHUB (\url{https://old.datahub.io/}) and \textit{LiveSchema} (\url{http://liveschema.eu/}) }
\vspace{-0.1cm}
\end{itemize}

%
The overall \itelos\ process is articulated in four phases. 
The \textit{Inception} phase takes as input the purpose and collects the ontologies and datasets needed to build the target EG. During this phase, 
the functional requirements are encoded into a list of CQs which are then matched to the input datasets and ontologies thus implementing the activities of \textit{Ontology Collection} and \textit{Dataset Collection}, as from Figure \ref{fig:pipeline}. 
The main objective of the \textit{Modeling} phase is to build the most suitable model of the ETG to be used as the schema of the final EG, which in Figure \ref{fig:pipeline} is called the \textit{ETG model}. In practice, the ETG model includes all the etypes and properties needed to represent the information required by each CQ, possibly extended by extra etypes and properties suggested by the datasets. The \textit{Dataset Selection} activity finalizes the selection of the datasets selecting whenever possible, the ones which are most popular.
The goal of \textit{Knowledge Alignment} is to enhance the sharability of the final EG, building in turn a shareable ETG that fits in the best possible way the datasets to be integrated. The input ETG model is itself a possible solution. However the set of reference schemas provides more possibilities, in terms of etypes and properties available, that can be adopted to implement a final ETG (called ETG in Figure \ref{fig:pipeline}) easier to share and reuse. Here again preference is given to those reference schemas which are most popular. 
The \textit{Dataset Cleaning} activity performs the final cleaning of the datasets consistently with the ETG, trying to minimizing it and concentrating the preprocessing on those datasets which are less popular. As described in detail in Section \ref{sec6_share}, this may require building an ETG which is an adaptation of the input reference schemas, e.g., in the selected etypes, properties, data types and formats. 

The last phase is \textit{Data Integration}. The objective is to build the EG, what we call \textit{EG Generation}, integrating the ETG with the data resources. 
 To do that, the ETG and datasets are provided in input to a specific data mapping tool, called \textit{KarmaLinker}, which consists of the \textit{Karma} data integration tool \cite{knoblock2015exploiting} extended to perform Natural Language Processing on short sentences (i.e., what we usually call \textit{the language of data}) \cite{bella2020exploring}. The process is described in some detail in \cite{2020-Stratified-DI}.
The first activity in this phase maps the data to the etypes and properties of the ETG. The following step is the generation of the entities that are then matched and, whenever they are discovered to be different representations of the same real world entity, merged.
These activities are fully supported by Karmalinker.
The above process is iteratively executed over the list of selected datasets. The process concludes with the export of the EG into an RDF file.

The key observation is that the \itelos\ is implemented as two separate sub-processes, executed in parallel within each phase, one operating on reference schemas, the other on the input data (blue and green boxes in Figure \ref{fig:pipeline}).  During this process, the initial purpose keeps evolving building the bridge between CQs, datasets and reference schemas. To enforce the convergence of this process, and also to avoid making costly mistakes, each phase ends with an evaluation activity (\textit{Eval} boxes in Figure \ref{fig:pipeline}). The details of how the evaluation is performed is out of the main scope of the paper. Here it is worth noticing that, within each phase, the evaluation aims to verify that the target of that phase is met, namely: aligning CQs with datasets and ontologies in phase 1, thus maximizing reusability; aligning the ETG model with the datasets in phase 2, thus guaranteeing the success of the project; and aligning ETG and ontologies in phase 3, thus maximizing sharability. The evaluation in phase 4 has the goal of checking that the final EG satisfies the requirements specified by the purpose. 
\vspace{-0.1cm}

\section{Enhancing data reuse} 
\label{sec5_reuse}

Reusability is enhanced during the phases of inception and modeling, whose main goal is to progressively transform the specifications from the purpose into the ETG Model. This process happens according to the following steps:

\begin{enumerate}
    \item generation of a list of natural language sentences, each informally defining a CQ, as implicitly or explicitly implied by the purpose; 
    \item generation of a list of relevant etypes and corresponding properties, which formalize the informal content of CQs, as from the previous step;
    \item selection of the datasets whose schema informally matches the CQs, as from the previous step;
     \item generation of a list of etypes with associated properties, from the selected datasets, which match the etypes and properties from the CQs;
     \item construction of the ETG model;
\end{enumerate}
 \noindent
 Steps 1-4 happen during inception, while step 5 happens during the modeling phase. 
 The key observation is that, starting from an analysis of the etypes and properties inside CQs, the two types of resources involved (i.e., ontologies, datasets) are handled through a series of three iterative executions, each corresponding to a specific category, following a decreasing level of reusability. The categories are defined as follows:

\vspace{0.2cm}
\noindent
\textit{Common}: this category involves resources associated with aspects that are common to all domains, also outside the domain of interest. Usually, these resources correspond to abstract etypes specified in \textit{upper level ontologies} \cite{DOLCE}, e.g., \textit{person}, \textit{organization}, \textit{event}, \textit{location}, and/or to etypes from very common domains, usually needed in most applications, e.g., \textit{Space} and \textit{Time}. 
The data that are found in Open Data sites as well the ontologies which can be found in the repositories mentioned above are examples of common resources.
   
\vspace{0.2cm}
\noindent 
  \textit{Core}: this category involves resources associated with the more core aspects of the domain under consideration. They carry information about the most important aspects considered by the purpose, information without which it would be impossible to develop the EG. Consider for instance  the following purpose:
    
  \vspace{0.1cm}
  \noindent
        ``\textit{There is a need to monitor Covid-19 data in the Trentino region (Italy), to understand the  diffusion  of  the  virus and  the  social  restriction  caused  by  the  virus,  with  the possibility to identify new outbreaks}".\footnote{This is a small example extracted from the project which built a KG, following the \itelos\ methodology, on \textit{``Integration of medical data on Covid-19"} developed by Antony, N., Gotca, D., Jyate, M., Donini, L. The complete material and description can be found at the URL \url{https://github.com/UNITN-KDI-2020/COVID-data-integration}. }

\vspace{0.1cm}
\noindent
    In this example, core resources could be those data values reporting the number of Covid-19 infections in the specified region. Examples of common resources are the data of certain domains, e.g., public sector facilities (e.g., hospitals, transportation, education),  domain specific ontologies that can be found again in the repositories above, as well as domain specific standards (e.g., Health, interoperability standards of various types). In general, data are harder to find than ontologies, in particular when they are about the private sector, where they carry economic value and are often collected by private bodies.

\vspace{0.2cm}
\noindent
\textit{Contextual}: this last category involves resources that carry specific, possibly unique, information from the domain of interest. These are the resources whose main goal is to create added value. If core resources are necessary for a meaningful application, contextual resources are the ones which can make the difference with respect to the competitors. In the above example, examples of contextual resources can be those data describing the type of social restrictions adopted to contrast the virus. At the schema level, contextual etypes and properties are those which differentiate the ontologies which, while covering the same domain, actually present major differences. Data level contextual resources are usually not trivial to find, given their specificity and intrinsic. In various applications we have developed in the past, this type of data have turned out to be a new set of resources those had to be generated on purpose for the application under development, in some cases while in production. 

The overall conclusive observation is that the availability of resources, and of data in particular, decreases from common to contextual. On top of this and because of this, as part of the \itelos\ strategy, a  decreasing effort is made, moving from common to core to contextual data, in maintaining high the level of reusability and sharability, thus concentrating the preprocessing costs in the latter categories.

\section{Enhancing data sharability} 
\label{sec6_share}

The knowledge alignment phase aims to enhance sharability, by aligning and possibly modifying the ETG model to take into account the etypes and properties coming from the reference ontologies. The key observation is that the alignment mainly concerns the common and, possibly, the core types with much smaller expectations on contextual etypes. Notice how the alignment with the most suitable ontology will enable the reuse of data, at least for what concerns common and sometimes core etypes. As an example, the selection of Google GTFS or FOAF as reference ontologies ensures the availability of a huge amount of data, a lot of which are open data. This type of decision should be made during inception; if delayed up to here, it might generate backtracking. 

We align the ETG model with the reference ontologies, by adapting the \textit{Entity Type Recognition (ETR)} process proposed in \cite{2020-KR}. This process happens in three steps as follows:

\vspace{0.1cm}
\noindent
\textit{Step 1: ontologies selection}. This step aims at selecting the set of reference ontologies that best fit the ETG model. As from above the first step is to rank the ontologies, as selected during the Inception phase, based on their popularity. Then, moving from top to down in the list, as from \cite{2020-KR}, this selection step occurs by measuring each reference ontology according to three metrics, which allow:
    \begin{itemize}
    \vspace{-0.2cm}
        \item to identify how many etypes of the reference ontologies are in common with those defined in the ETG model,  and
        \item to measure a property sharability value for each ontology etype, indicating how many properties are shared with the ETG model etypes. 
    \end{itemize}
\noindent The output of this first step is a set of selected ontologies, which best cover the ETG model, and that have been verified fitting the dataset's schema, at both etypes and etype properties levels.


\vspace{0.1cm}
\noindent 
\textit{Step 2: Entity Type Recognition}(ETR). The main goal here is to predict, for each etype of the ETG model, which etype of the input ontologies, analyzed one at the time, best fits the ETG. In practice, the ETG model's etypes are used as labels of the classification task and, as mentioned in \cite{2020-KR}, the execution exploits techniques that are very similar to those used in ontology matching (see, e.g., \cite{2012-Giunchiglia3}). The final result is a vector of prediction values, returning a similarity score between the ETG model's etypes and the selected ontology etypes. 

\vspace{0.1cm}
\noindent
\textit{Step 3: ETG generation}. This step identifies, by using the prediction vector produced in the previous step, those etypes and properties from the ontologies which will be added to the final version of the ETG. Notice how this must be done while preserving the mapping with the datasets' schemas. 
The distinction among common, core and contextual etypes and properties plays an important role in this phase and can be articulated as follows:
\begin{itemize}
\vspace{-0.1cm}
    \item The common etypes should be adopted from the reference ontology, in percentage as close as possible to 100\%. This usually results in an enrichment of the top level of the ETG model by adding those top level etypes (e.g., \textit{thing}, \textit{product}, \textit{event}, \textit{location}) that usually no  developer considers, because too abstract, but which are fundamental for building a highly shareable ETG where all properties are positioned in the right place. This also allows for an alignment of those \textit{common isolates} (see Section \ref{sec5_reuse}) for which usually a lot of (open) data are publicly available (e.g., \textit{street});
    \item The core etypes are tentatively treated in the same way as common etypes, but the results highly depend on the ontologies available. Think for instance of the GTFS example above;
    \item Contextual etypes and, in particular, contextual properties are mainly used to select among ontologies, the reason being that, they allow distinguishing the most suitable among a set of ontologies about the same domain \cite{2020-KR}.
    \vspace{-0.1cm}
\end{itemize}


\section{Case studies} 
\label{sec8_case}

The specification of \itelos\ is in its early phases, in particular in terms of tool support. However, a lot of work has been dedicated to the refinement of the single steps of the overall approach and on their extensive evaluation. In particular, \itelos\ has been validated during the past four Academic Years as part of the Knowledge and Data Integration (KDI) class, a six credit course of the Master Degree in Computer Science of the University of Trento.\footnote{\url{http://knowdive.disi.unitn.it/teaching/kdi/} contains the material used during the last two editions of the course. This material consists of theoretical and practical lectures, as well as demos of the tools to be used, some of which have been mentioned above.} 
During this class, 2-5 students per group, must generate an EG using the pipeline above starting from a high level problem specification. The overall project has an elapsed time of fourteen weeks during which students have to work intensely. We estimate the overall effort each group puts into building an EG in around 4-8 person-months, depending on the case. 

As of today we have piloted around 30 projects and 90 evaluations of the \itelos\ methodology as a whole.
The details of this work cannot be reported here for lack of space. However the results, restricted to the first three years are described in some detail in \cite{KD-2022-Bocca}. We report below the most relevant answers provided by the students, which were asked to fill a very detailed questionnaire containing a set of qualitative questions about the methodology. 

\begin{itemize}
\item \textit{(Strength)} the step by step, precisely articulated, \itelos\ process is easy to follow;
\item \textit{(Strength)} the stepwise iterative evaluation process supports well the refinement of the Entity Graph; 
\item \textit{(Weakness)} A wrong decision made in the early phases is quite difficult to remedy, with this possibility being very high during the inception phase;
\item \textit{(Weakness)} The work between the schema and the data layer is unbalanced in favour of the second, in particular during the \textit{informal modeling} phase. This complicates the synchronization of the work with the possibility of misalignments, mainly because of misunderstandings, that have to be handled very carefully by the project manager.
\end{itemize}

\section{Conclusions} 
\label{sec9_conclusion}

In this paper we have introduced \itelos, a novel methodology whose ultimate goal is to implement a \textit{circular} development process. By this we mean that the goal of \itelos\ is to enable the development of EGs via the  \textit{reuse} of already existing EGs and ETGs, while being simultaneously developed to be later easily \textit{reused} by other applications to come. 
%

\bibliographystyle{splncs04}
\bibliography{KnowDivePubs}

\end{document}